\documentclass[11pt]{report}
\usepackage{upennstyle} 

\DeclareMathOperator{\sgn}{sgn}
\DeclareMathOperator{\atan}{atan}

\usepackage[nonamebreak,round]{natbib}
\usepackage{amsmath}
\usepackage{amssymb}
\usepackage{mathtools}
\usepackage[utf8]{inputenc}
\usepackage[T1]{fontenc}
\usepackage[thinc]{esdiff}
\usepackage{caption}
\usepackage{subcaption}
\title{GAZE-BASED LEARNING FROM DEMONSTRATION IN SURGICAL ROBOTICS}
\author{Daniel Chen, Jaden Ingleton, Trent Suzuki, Sayem Zaman}
\gradgroup{2210 - Eye Gaze Tracker} 
\date{2022} 
\supervisor{Alaa Eldin Abdelaal}
\supervisortitle{Vanier Scholar and PhD candidate
UBC Robotics and Control Lab}
\gradchair{Graduate Group Chairperson's Typed Name, Full Faculty Title}
\committee{Dylan Gunn, Project Lab Director}
\committee{Miti Isbasescu, Head of Software Systems}
\committee{Bernhard Zender, Student Support} 

\abstract{Surgical robotics is a rising field in medical technology and advanced robotics. Robot assisted surgery, or robotic surgery, allows surgeons to perform complicated surgical tasks with more precision, automation, and flexibility than is possible for traditional surgical approaches. The main type of robot assisted surgery is minimally invasive surgery, which could be automated and result in a faster healing time for the patient. The surgical robot we are particularly interested in is the da Vinci surgical system, which is developed and manufactured by Intuitive Surgical. In the current iteration of the system, the endoscopic camera arm on the da Vinci robot has to be manually controlled and calibrated by the surgeon during a surgical task, which interrupts the flow of the operation. The main goal of this capstone project is to automate the motion of the camera arm using a probabilistic model based on surgeon’s eye gaze data and da Vinci robot kinematic data. We believe that this capstone project will not only alleviate surgeons’ physical and mental load during surgical tasks, but also push the advancement of surgical robotics.} 

\begin{document}
\maketitle 
\setcounter{page}{2}



\makeacknowledgement 
The da Vinci surgical system is one of the most advanced surgical robots developed in the past decade. Camera control in the da Vinci surgical system involves pressing a foot-pedal to transition from tool arm control to manual camera arm control. For long-duration surgeries, multiple transitions between tool arms and the camera arm can increase cognitive load on surgeons.

Alaa Eldin Abdelaal, a PhD candidate at the UBC Robotics and Control Laboratory, is interested in understanding the role of eye gaze in autonomous camera control. The goal of this project is to automate the camera control using surgeon eye gaze and tool arm kinematic data. This will allow the surgeon to focus solely on controlling the tool arms while the camera arm autonomously adjusts itself to a comfortable position and orientation during a surgical task. 

Over the span of this project, we developed a model for camera placement prediction. With the addition of eye gaze and orientation data, our machine learning models were able to successfully predict the camera placement with precision on the order of millimeters. As a next step for the project, we recommend qualitative and quantitative evaluation of our model for autonomous camera control by expert surgeons using the criteria outlined in this report.

\makeabstract
\tableofcontents
\newpage
\clearpage \phantomsection \addcontentsline{toc}{chapter}{LIST OF TABLES} \listoftables 

\clearpage \phantomsection \addcontentsline{toc}{chapter}{LIST OF FIGURES} \listoffigures 


\begin{mainf} 
\chpt{Commonly used Terms}
\begin{center}
\begin{table}[ht]
        \centering
        \scalebox{1}{
        \begin{tabular}{l l}
        \hline
         \textbf{Term} & \textbf{Description}\\
        \hline
        End Effector & The device at the end of a robotic arm \\ & designed to interact with the environment\\\\

        Pose & A position and orientation pair\\\\

        Revolute Joint & Joint which rotates about one axis\\\\
 
        Prismatic Joint & joint which translates along one axis\\\\

        Remote Center of Motion (RCM) & Centre of rotation fixed at a point \\ &where no joint exists\\\\

        Robot Operating System (ROS) & Set of software libraries and tools to support \\ &robot applications\\\\

        Patient Side Manipulator (PSM) & Arm used to hold tools\\\\

        Endoscope Camera Manipulator (ECM) & Arm mounted with endoscope camera\\\\

        Master Tool Manipulator (MTM) & Controller for patient side arms\\\\

        Setup Joint (SUJ) & Links to hold arms in place\\\\
        
        Gaussian Mixture Model (GMM) & A type of probabilistic model\\\\
        
        Gaussian Mixture Regression (GMR) & Regression function for GMM\\\\
        
        Learning from Demonstration (LfD) & A machine learning method\\\\
        
        Bayesian Information Criterion (BIC) & A evaluation method for \\& machine learning models\\\\%

        \end{tabular}}
        \caption{Commonly Used Terms}
        \label{tab:Performance Measures for Proportional}
    \end{table}
\end{center}
\chpt{Introduction}
\section{Information about the Sponsor}
Alaa Eldin Abdelaal is a Vanier Scholar and PhD candidate at the Electrical and Computer Engineering Department at the University of British Columbia. He is currently working as a research assistant at the Robotics and Control Laboratory and a visiting graduate scholar at the Computational Interaction and Robotics Lab (CIRL) at the Department of Computer Science at Johns Hopkins University, co-advised by Dr. Tim Salcudean and Dr. Gregory Hager. His research focuses primarily on surgical robotics, Human-Robot Interaction, and Automation. Through this project, he hopes to gain a deeper understanding of the role of eye gaze in robot assisted surgery. 

\section{Background and Significance of the Project}
The goal of robot-assisted surgery is to enhance a surgeon's capabilities by providing the surgeon with an experience that is more convenient and efficient than that of conventional (non-assisted) surgery.

The majority of the most popular surgical robotics systems, including the da Vinci and ZEUS systems, employ three robotic arms: Two patient-side manipulators (PSMs), which the surgeon can use to grab and manipulate tools and objects as they would with their hands during a conventional surgery; and one camera module, which provides the surgeon with a stereoscopic 3D view of the environment.

The input mechanism for control of the camera arm position is still an active research topic. In the current iteration of the da Vinci system, the surgeon can manipulate the angle and position of the camera by first pressing a foot pedal, then while holding the pedal down, moving his or her arms to translate and rotate the camera arm. There is a problem with this type of control scheme: When the surgeon wishes to manipulate the angle and position of the camera, they must temporarily relinquish control of the patient-side manipulators. This means that at any instant in time, the surgeon can only move either the patient-side manipulators or the camera module. This approach is inefficient and unnatural compared to conventional surgery. Voice-controlled camera arms have also been implemented \citep{onathan}. For example the camera arms can be maneuvered by simple verbal commands, such as “up”, “down”, “left”, “right”, “stop”, etc. This approach enables a single surgeon to operate without assistance by automatically moving the camera arm according to the surgeon’s commands with the limitation of camera motion to a fixed 2D plane. While this system managed to lower the occurrence of interruptions during a surgery, the surgeon may still occasionally be required to pause the surgery and actively control the camera arm.

The role of eye gaze in camera control has also been explored. In one paper by Ali et al., the camera was made to follow the center of the surgeon’s vision using eye-gaze tracking \citep{ali}. This system, however,  did not account for other environmental factors such as the positions of the robot arms or features within the surgeon’s field of view. Another approach \citep{king} instructed the camera arm to ensure that the surgical tools remain visible at all times by calculating their centroid on the camera feed using image processing techniques and centering the camera view on that position. Although this control scheme succeeded in its goals, the authors recognized that because of the limitations of the algorithm, this system would not be comprehensive enough for use during an actual surgical operation.

Integrating the ideas from previous work, we proposed a new approach to camera control based on a combination of inputs: The surgeon’s eye-gaze position and the robot kinematic data. This method would use machine learning to infer the best position and angle in which to place the camera given the current state of the surgical environment. If successful, this new method could help streamline the surgical process by providing a smooth and continuous view for the surgeon, enabling them to operate without interruptions. We also believe that because of the multivariate input data, the model might generalize well to a variety of tasks. By determining the relationships between the chosen input variables and the camera position, this project could provide insight into the key parameters governing the optimal camera position, and serve as a stepping stone to a more generalized model.

\section{Statement of the Problem and Project Objectives}
Currently, surgeons control the surgical camera and the robot arms using two hand-held controllers. To switch between control modes, the surgeon has to relinquish control of the arms, which interrupts the flow of the operation. The main goal of this research is to automate the motion of the endoscopic camera arm using a probabilistic model based on eye gaze data and robot kinematic data. The model will take in the expert surgeons’ eye gaze data and robot kinematic data of the surgical scene to learn how to automate the camera movement. 

\section{Scope and Limitations}
The inputs we considered for our machine learning model are: PSM positions and orientations, and the location of surgeon eye-gaze. Additional inputs such as the ECM camera feed, type of surgery, and surgeon pupil diameter were initially considered, but were not included due to time constraints.

The choice of a machine learning model type, implementation of this model, and verification of the results all fall within the project scope. The outputs of this model were to be validated before being sent to the da Vinci, meaning the output is guaranteed to be a physically reachable and non-singular configuration. Other invalid configurations such as those where the ECM and PSM(s) collide, or where the ECM collides with some object (such as a patient) in the environment were not to be considered within the project scope.

\chpt{Discussion} \label{discussion}
\section{Approach and System Overview}
Based on our research into the current literature for autonomous camera control in surgical robotics, we decided to create a model to predict the camera pose using a technique called Learning from Demonstration. This approach aims to extract task parameters from data collected during expert demonstrations, rather than explicitly commanding the robot to perform certain motions. We chose Learning from Demonstration because it has been previously employed in the context of surgical robotics, and it has been shown to require a relatively small amount of data compared to other machine learning techniques \citep{LfD}. Several other research groups, most notably \citep{transfering} have successfully used Learning from Demonstration to predict camera positions based on cartesian positions of the robot arms during a surgical task. Our aim was to expand on this result by adding the surgeon’s point of gaze and the orientation of each arm. More specifically, we planned to use the robot tool end effector kinematic data and the surgeon eye gaze to predict the most probable ECM position and orientation. 

\begin{figure}[H]
    \centering
    \includegraphics[scale=0.5]{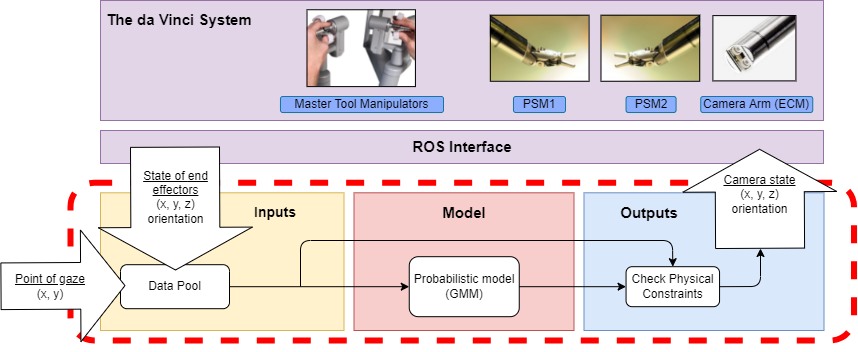}
    \caption{System Level Diagram}
    \label{fig:SystemLevelDiagram}
\end{figure}

The system level diagram for the project is shown in Figure \ref{fig:SystemLevelDiagram}. We will cover each component of this diagram in the following subsections. Although we implemented our system using the da Vinci, the principles of camera control generalize to any robotic surgical system. We used the da Vinci robot interface to collect robot kinematic data, and used a pre-built eye gaze tracker to obtain the surgeon’s point of gaze. We fed this data into our machine learning model which generated a prediction for the position of the endoscopic camera. Once we had this prediction, we validated the position as reachable within the current context before commanding the da Vinci robot to move to this location.

\section{Data Collection}
\label{data Collection System}
To obtain the position and orientation data from the da Vinci surgical system, we used the da Vinci Research Kit (dVRK), an open source ROS based control system developed at Johns Hopkins University. The dVRK consists of electronic controllers and firmware that provide a real-time interface with the da Vinci robot. See Section \ref{Hardware and Software} for more information on the dVRK.

To collect surgeon eye gaze data, we used a custom eye gaze tracker developed by a previous MASc student at the UBC Robotics and Control Laboratory, Irene Tong.
\begin{figure}[H]
    \centering
    \begin{subfigure}[b]{0.4\textwidth}
    \includegraphics[scale=0.5]{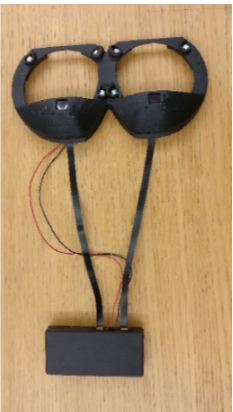}
    \caption{Eye gaze tracker headset}
    \label{fig:EyeGazeTracker2}
    \end{subfigure}
    \begin{subfigure}[b]{0.4\textwidth}
    \includegraphics[scale=0.5]{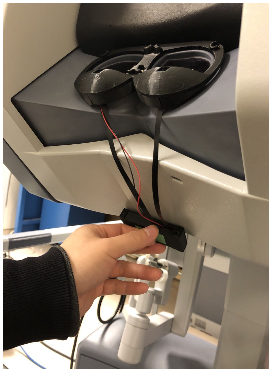}
    \caption{Eye gaze tracker placed in surgeon console}
    \label{fig:EyeGazeTracker}
    \end{subfigure}
\end{figure}
The eye gaze tracker is supported by an eye gaze tracking software application, GazeTrackGUI. By shining infrared light on the surgeon’s eye and measuring the reflected glints, the device can interpret where the surgeon is looking on the stereoscopic 3D view provided by the da Vinci surgeon console. This data is reported as left and right points of gaze (POG) in xy coordinates, pupil diameters, reflected infrared glint locations and pupil locations \citep{Tong_2017}.
\begin{figure}[H]
    \centering
    \includegraphics[scale=0.9]{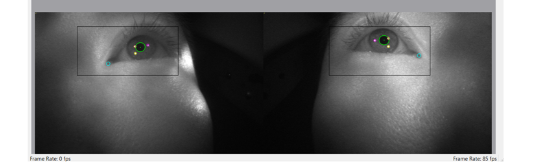}
    \caption{Eye gaze tracker software (screenshot)}
    \label{fig:EyeGaze Software}
\end{figure}
Because both data sources require a high CPU load, the two components were run on separate PCs as shown in Figure \ref{fig:Data Collection}:

\begin{figure}[H]
    \centering
    \includegraphics[scale=1]{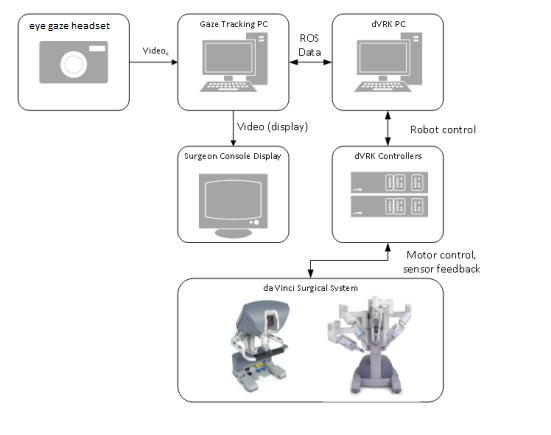}
    \caption{Communication channels between dVRK PC, eye gaze software and da Vinci \citep{Tong_2017}}
    \label{fig:Data Collection}
\end{figure}
To synchronize the kinematic data with the eye gaze data, the surgeon gaze data from the eye gaze PC was sent to the dVRK PC via TCP/IP connection, where it was then published to ROS topics. We were able to create a synchronized pool of kinematic and eye gaze data by ensuring messages are being published to the ROS topic at the same rate. The kinematic data of the robot arms, the endoscopic camera view from the ECM, and the eye gaze data are all obtained by subscribing to their respective ROS topics and stored in a CSV file with the corresponding time stamp. 
The data collection system is designed to run in the background while a surgical task is being performed and captures all relevant data during the task.

The final contents of the data pool are summarized in Table \ref{tab:Performance Measures for Proportional}. 
\begin{table}[ht]
        \centering
        \scalebox{0.85}{
        \begin{tabular}{|l||l|}
        \hline
         \textbf{Data} & \textbf{Format}\\
        \hline
        Timestamp & Epoch Time\\
        \hline
        MTM joint angles & [Joint1, Joint2, Joint3, Joint4, Joint5, Joint6, Joint7, Jaw]\\
        \hline
        MTM position & [x, y, z] cartesian\\
        \hline
        MTM orientation & $3 \times 3$ Orientation Matrix\\
        \hline
        ECM joint angles & [Joint1, Joint2, Joint3, Joint4]\\
        \hline
        PSM1, PSM3 joint angles & [Joint1, Joint2, Joint3, Joint4, Joint5, Joint6, Jaw]\\
        \hline
        PSM1, PSM3, ECM position & [x, y, z] cartesian  (see Section \ref{Frame Transform to WC})\\
        \hline
        PSM1, PSM3, ECM orientation & $3 \times 3$ Orientation Matrix (see Section \ref{Frame Transform to WC})\\
        \hline
        Left, Right, Best Point of Gaze (POG) & [LPOGx, LPOGy, RPOGx, RPOGy, BPOGx, BPOGy]\\
        \hline
       Left, Right Pupil Diameters & [Left Pupil Diameter, Right Pupil Diameter]\\%
        \hline
        \end{tabular}}
        \caption{Data Collection Format}
        \label{tab:Data Collection Format}
    \end{table}

\section{Gaussian Mixture Models}
\subsection{Overview}
After collecting our input data, the data was fed into the Learning from Demonstration model. To implement Learning from Demonstration, we were required to choose a machine learning model. We chose to use a Gaussian Mixture Model, the same model type used by \citep{transfering}, which had been used to predict camera positions based on cartesian positions of the robot arms during a surgical task. Our aim was to expand the Gaussian Mixture Model by adding the surgeon’s point of gaze and the orientation of each arm.

Gaussian Mixture Models are a probabilistic method of generating outputs. This approach assumes all the data points are generated from a mixture of a fixed number of multivariate Gaussian probability distributions. Each Gaussian probability distribution has a mean, a covariance matrix, and prior probability. To determine these parameters, we used the Expectation Maximization algorithm \citep{sylvain}, an iterative process that trains the model to fit the data from expert demonstrations. For each dataset, we trained models with different numbers of Gaussians, beginning from one and ending as high as our computer hardware would allow. We then selected the best-performing model using the Bayesian Information Criterion (BIC). The parameters for the model with the lowest BIC score, i.e. with the best performance, were exported for deployment.

After a model is trained, the desired output can be obtained using Gaussian Mixture Regression (GMR), which determines the most probable set of outputs from the set of Gaussian clusters that best represents the entire dataset. In our case, Gaussian Mixture Regression was used to obtain the most probable endoscopic camera position and orientation for a given set of inputs from the trained Gaussian Mixture Model. 

\subsection{Code Implementation}

To implement the Gaussian Mixture Models in code, we chose to use the \\ \underline{\href{https://scikit-learn.org/stable/modules/generated/sklearn.mixture.GaussianMixture.html}{Python Scikit Learn implementation}}. This offered a well-documented, reliable setup for training models. On top of this framework, we built a training interface that made it simple to add or remove input and output dimensions, which was a crucial aspect of our approach of incrementally increasing model complexity. As for the Gaussian Mixture Regression, we implemented it by translating the Matlab code from the research by \cite{transfering} to python code. We also validated the results of the GMR with the result trained by the Matlab code and another \underline{\href{https://github.com/AlexanderFabisch/gmr}{external API}}.

\section{Model Training and Deployment Cycle}
To test our model and deployment, we first gathered training data using a simulated da Vinci robot in a simulation package software called CoppeliaSim. CoppeliaSim is a dynamic simulation software where each object/model can be individually controlled via an embedded script or a ROS node environment.

In the CoppeliaSim da Vinci robot simulation (Figure \ref{fig:CoppeliaSim}), we moved the robot arms in several trajectories, which included straight lines sinusoidal trajectories and a combination of both. Using such simple paths for the arms allowed us to easily evaluate model performance according to a known cost function. As the arms moved, we recorded the positions of the PSMs and ECM using the data collection system.

\begin{figure}[H]
    \centering
    \includegraphics[scale=0.7]{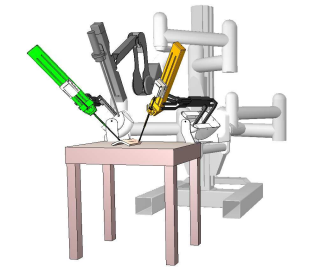}
    \caption{da Vinci robot setup in CoppeliaSim}
    \label{fig:CoppeliaSim}
\end{figure}

After collecting the data, we now had demonstrations from our simplified “surgical tasks”, and could begin training models. First, we split the dataset using an 80/20 training/validation split, assigning each data point randomly to either the training dataset or the validation dataset. We began with relatively simple models with 9 dimensional Gaussians: 6 input dimensions (2 PSMs [x,y,z]) and 3 output dimensions (ECM [x,y,z]). Once these models were trained, we tested them in simulation by commanding the robot arms to move in a similar manner to the trajectories used during training and predicted the camera position in real time. By measuring the error between the expected and actual positions of the ECM, we could quantitatively evaluate the model’s performance. After we were reasonably confident in the model’s accuracy in simulation, we moved to testing on the real da Vinci robot, using the same method of real-time prediction. Results from these tests are discussed in Section \ref{chpt:results1}. We developed a deployment pipeline to ensure the consistency of our system setup and accuracy of the data. A visual representation of the model deployment pipeline is shown in Figure \ref{fig:pipelinedeploy}.

The results from these tests allowed us to establish a baseline for model performance which we could use to isolate the contribution of additional input parameters. The first additional inputs we tried were the orientations of each robot arm. We chose to use quaternions to represent the orientation data because this form was more compact than orientation matrices and each component of the representation could be predicted independently, which is not possible with an orientation matrix.

We also planned to add the surgeon’s point of gaze as an input. However, due to time constraints we were unable to invite expert surgeons to try our system in the lab at UBC. Instead, we trained models on task data from expert surgeon demonstrations performed in 2016 using a dataset from Intuitive Surgical. This data was collected on the da Vinci SI, a slightly newer model than the one we have in the lab at UBC. Because of this, models we trained using this dataset were not compatible with the dVRK and could not be tested on the real robot. These results are discussed in Section \ref{chpt:results2}.

\section{Model Output Validation}

Since the LfD model chosen does not incorporate the physical limits of the ECM into its predictions, we need to verify the outputs are valid before sending them to the da Vinci. In addition, we would like to have a buffer region around the invalid points as a safeguard against errors in the da Vinci’s position sensors. In this section we review how points are classified as invalid or valid, how we obtain a mathematical model of such points, how a buffer is added, and the frame transformation necessary for implementation on the real robot.

A given pose is considered valid if it is inside the reachable workspace. Whether a pose puts the arm into a singular configuration need not be considered as the only two singularities of the ECM are outside of the joint limits, thus already considered invalid by the reachability condition. Figure~\ref{fig:ECM singularities} shows the singularities and the relevant range of motion of the ECM.

\begin{figure}[H]
    \centering
    \includegraphics[scale=1.2]{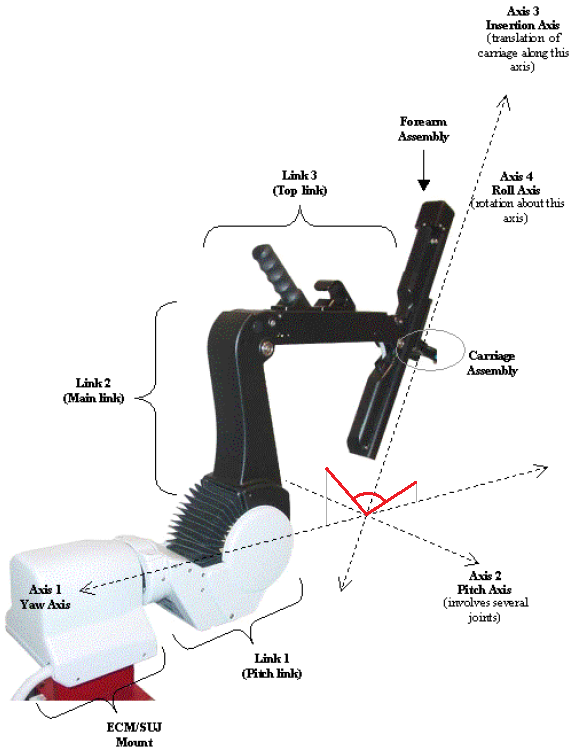}
    \caption{Singularities and the relevant range of motion of the ECM. \underline{\href{https://research.intusurg.com/images/2/26/User_Guide_DVRK_Jan_2019.pdf}{Source}}}
    \medskip
    \small
    The ECM with the joint axes overlaid. Singularities would occur if the ECM rotates about axis 2 until axis 3 becomes parallel or antiparallel with axis 1. This however is not possible to due the limits on joint 2, highlighted in red. Note axis 3 and axis 4 are coincident.
    \label{fig:ECM singularities}
\end{figure}

There are two approaches to determine whether a given pose is inside or outside of the reachable workspace - we could perform the inverse kinematics and check whether the joint angles are within their respective limits, or we could check that the position is inside of the reachable workspace and that the orientation is possible for that given position. Both approaches require equal work as both require obtaining the forward and inverse kinematic equations, however, the second approach lends itself far better to implementing a uniform buffer around the workspace boundary than the first approach. For these reasons the second approach was chosen.

To determine whether a point is within the reachable workspace, we need a description of the boundaries of the workspace in cartesian space. These can be calculated using the forward kinematics equations, which can be obtained using the Denavit-Hartenberg (DH) parameters of the ECM. An overview of the reachable workspace calculations can be found in Appendix \ref{reachableWS}, and a detailed description of DH parameters can be found in Appendix \ref{DH}.

With descriptions of the bounding surfaces, the validity of a point can be determined by finding the signed distance from each of the surfaces, and a uniform buffer can easily be added by requiring some nonzero signed distance from the bounding surfaces. The associated orientation is validated by using one of the inverse kinematic equations to obtain the angle of the fourth joint and ensuring it is within its limits plus the buffer added to it.

To ensure the model output validation would work for any configuration of setup joints, which may change between each surgery, the reachable workspace surfaces are described with respect to the ECM RCM. The transform between the ECM RCM frame and the frame the LfD model uses, the robot base frame, is queried from the da Vinci before each surgery from a ROS topic\footnote{The ROS topic that defines the ECM RCM with respect to the robot base frame is /SUJ/ECM/local/measured\_cp}.

\chpt{Results} \label{chpt:results}
\section{Results from Lab Data} \label{chpt:results1}
After iterating through several training cycles using the process described in Section \ref{discussion}, we arrived at a model composed of 46 Gaussians that takes in the PSM positions and outputs the ECM positions. As described in Section \ref{discussion}, we selected 46 Gaussians as the optimal number according to BIC from the models we trained, which contained between 1 to 60 Gaussian components. The dataset used to train this model consisted of a combination of linear and sinusoidal trajectories collected from the real robot, with a 80-20 train-validation split. Figures \ref{fig:resultfig1} - \ref{fig:resultfig6} were plotted by generating camera placement predictions along linear and sinusoidal trajectories with different frequencies and orientations. Notice that the predicted value versus expected value deviations often occur around the margins or boundaries of the curves. However, the average errors for most of the deployments lie within the millimeter region, which is acceptable considering that the position error of the da Vinci robot itself is reported as 5 cubic millimeters in the documentation.

\begin{figure}[H]
    \centering
    \includegraphics[scale=0.6]{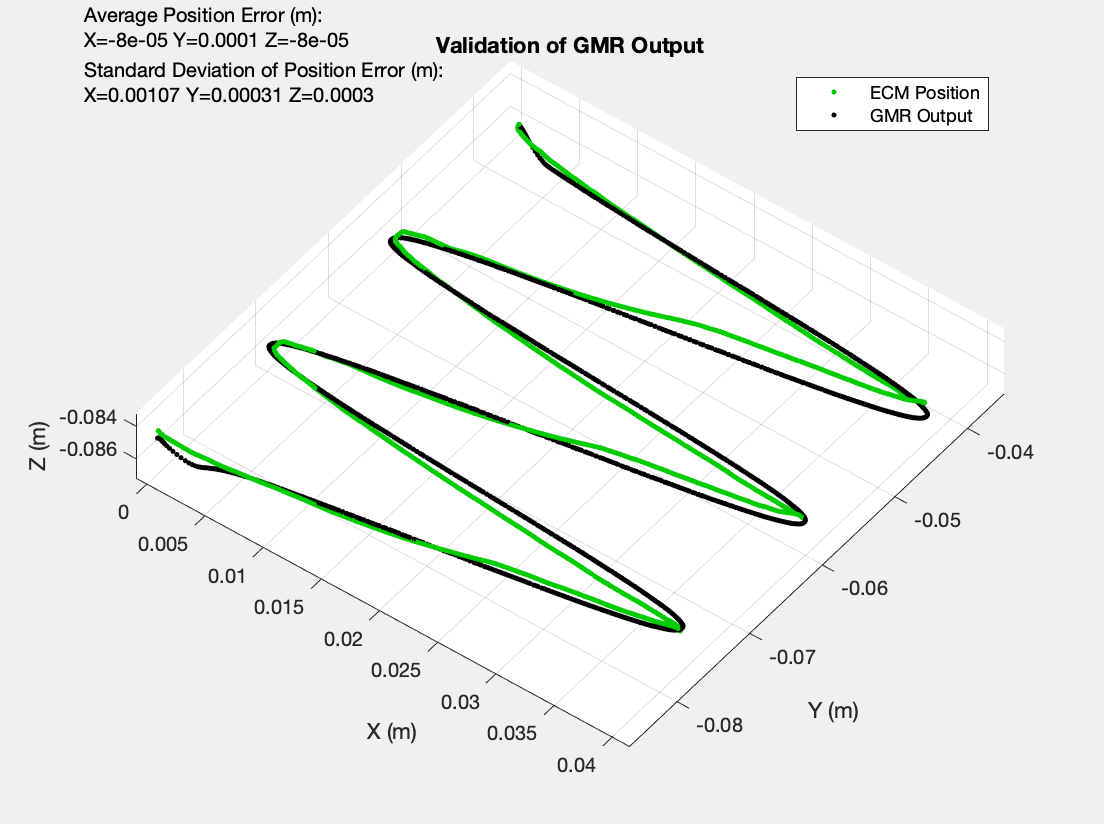}
    \caption{Sinusoidal trajectory in y direction expected vs. predicted}
\end{figure}

\begin{figure}[H]
    \centering
    \includegraphics[scale=0.6]{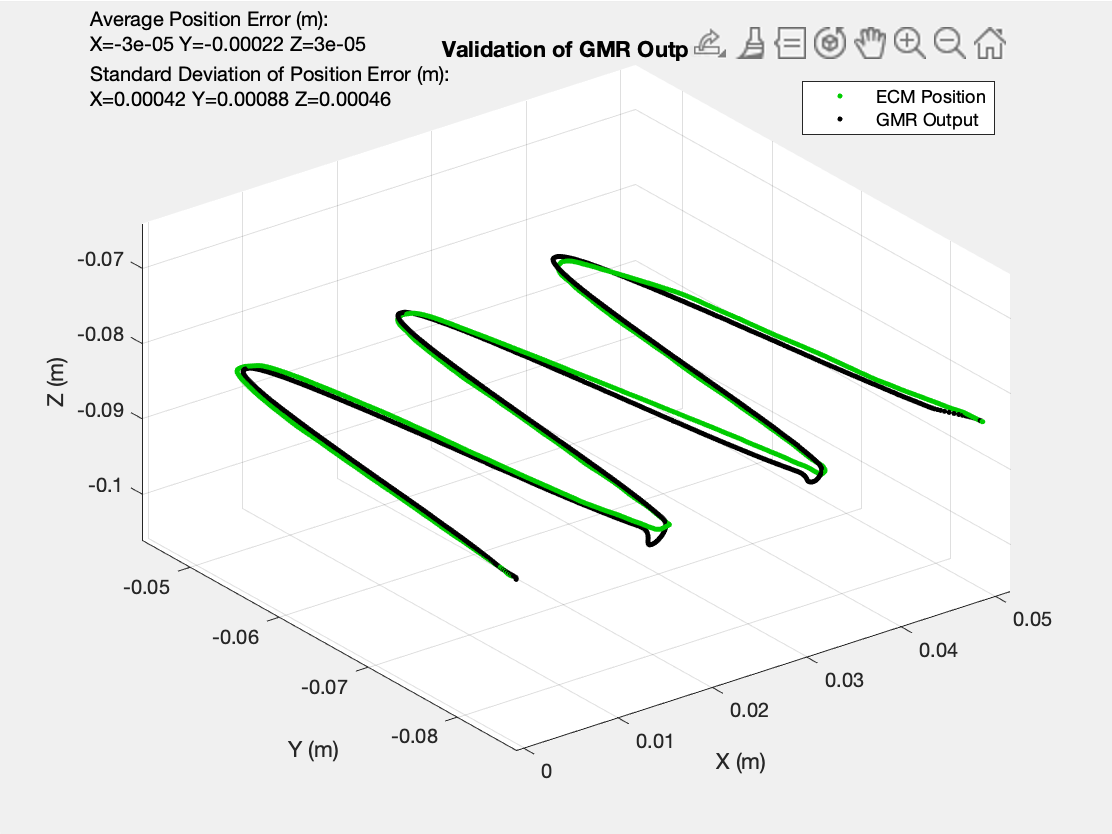}
    \caption{Sinusoidal trajectory in x direction expected vs. predicted}
    \label{fig:resultfig1}
\end{figure}

\begin{figure}[H]
    \centering
    \includegraphics[scale=0.6]{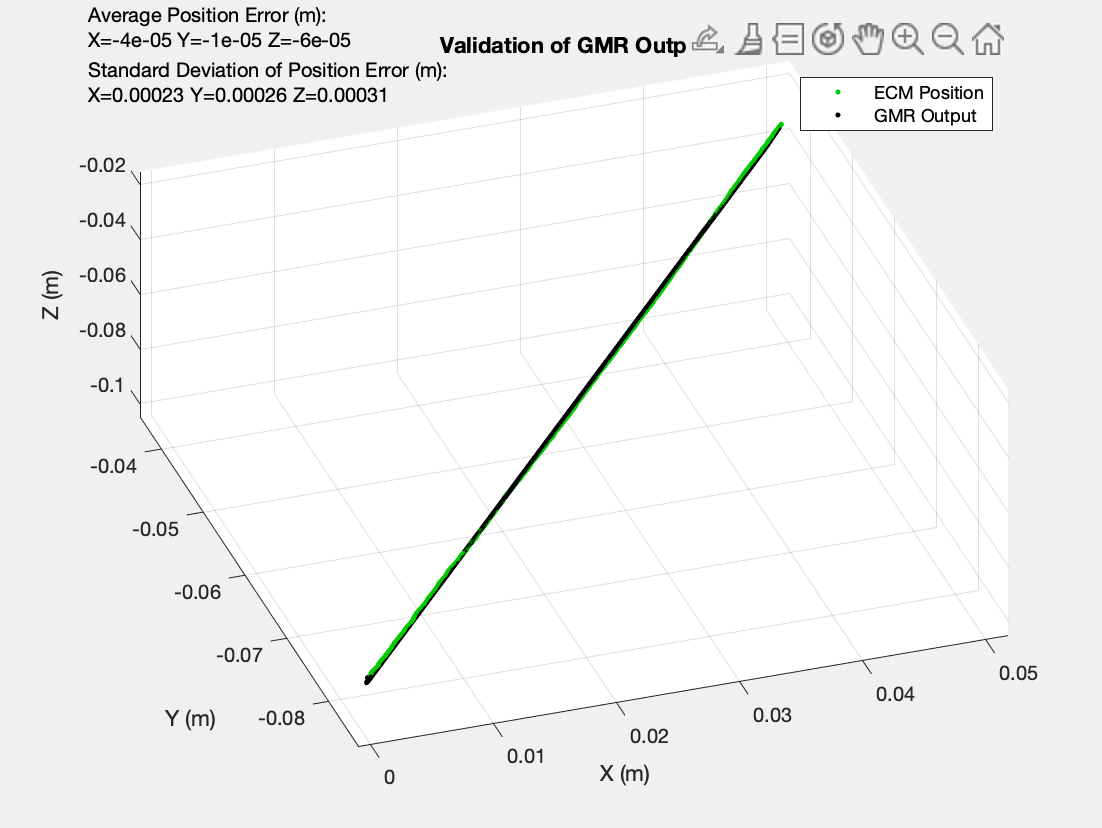}
    \caption{Linear trajectory in xyz direction expected vs. predicted}
\end{figure}

\begin{figure}[H]
    \centering
    \includegraphics[scale=0.6]{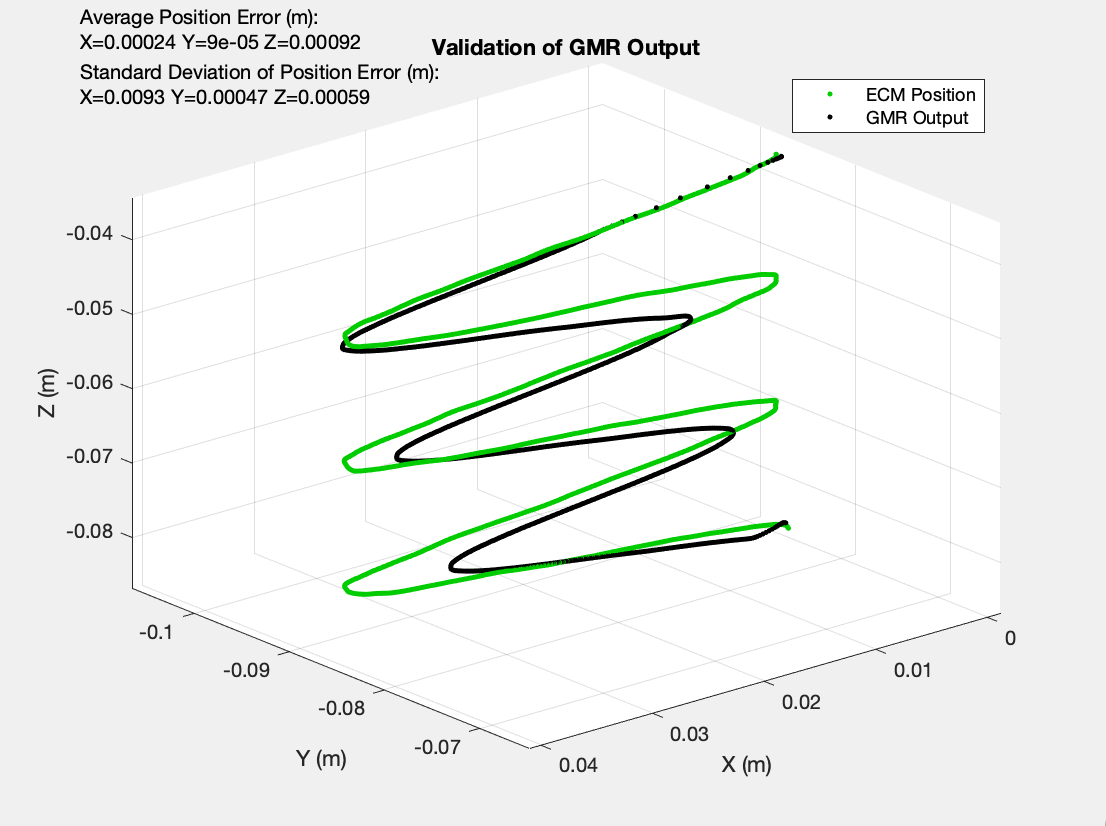}
    \caption{Sinusoidal trajectory in z direction expected vs. predicted}
\end{figure}

\begin{figure}[H]
    \centering
    \includegraphics[scale=0.6]{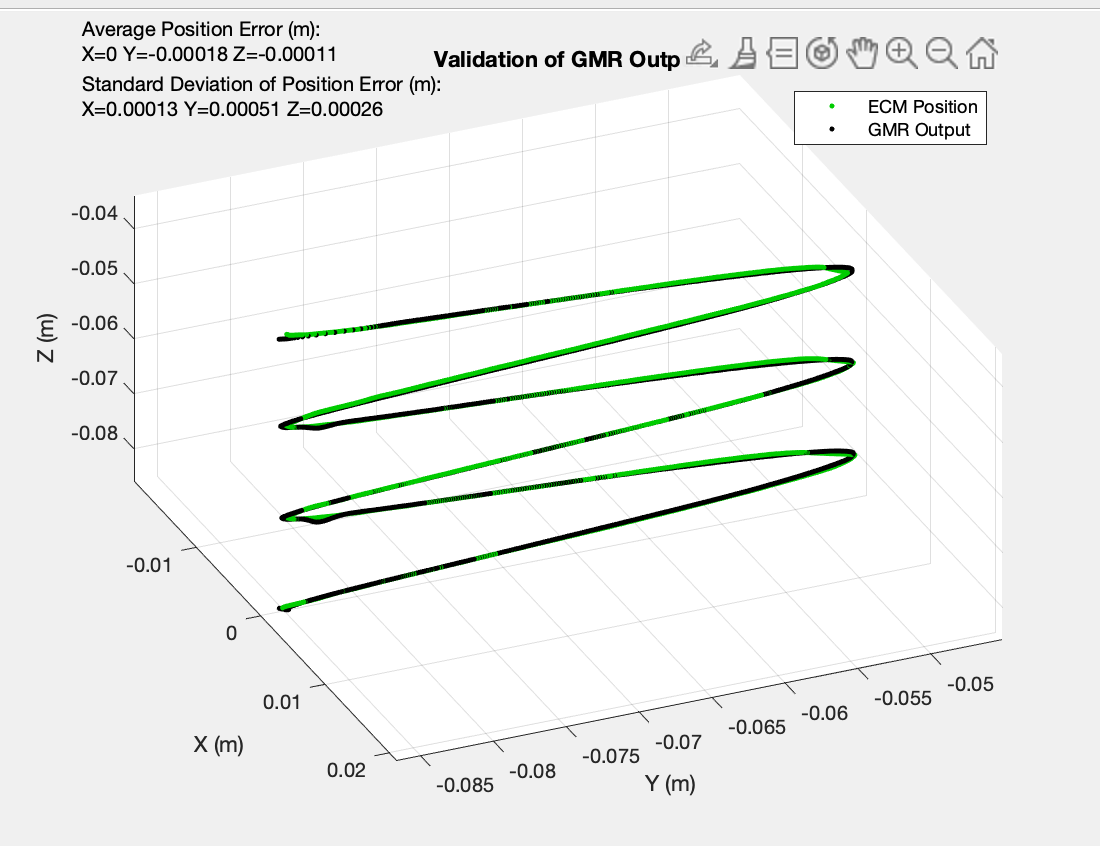}
    \caption{Sinusoidal trajectory in z direction expected vs. predicted}
\end{figure}

\begin{figure}[H]
    \centering
    \includegraphics[scale=0.6]{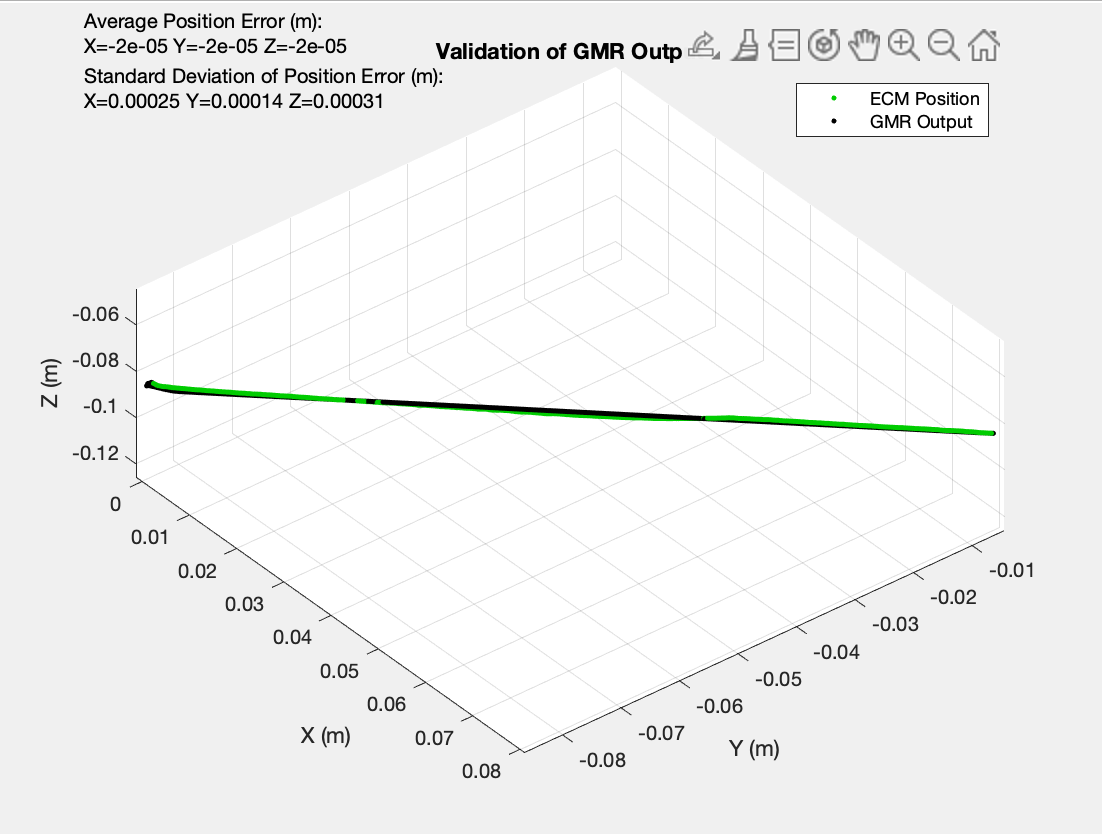}
    \caption{Linear trajectory in xy direction expected vs. predicted}
    \label{fig:resultfig6}
\end{figure}

\section{Results from Intuitive Surgical Data} \label{chpt:results2}

In addition to the model described above, we trained models using data from a 2016 dataset provided to us by Intuitive Surgical. This dataset contained recordings of surgical tasks performed by expert surgeons. Because this dataset was recorded on the da Vinci SI system, the foot pedal was used to operate the camera. This introduced a bias in our dataset because the training data was limited by the feature we wished to eliminate. After discussing with the project sponsor, we chose to filter the dataset to include only points when the foot pedal was pressed and the camera was moving. Using this dataset, we trained models on the data from one surgeon and one task, selected based on the data availability and range of motion throughout the task. Using an 80/20 training/validation split and up to 40 Gaussians, we selected the best-performing model by BIC. After training the model, we evaluated the performance on the remaining 20\% of the data by using GMR to generate a predicted camera position and comparing it to the actual camera position from the dataset.

\begin{figure}[H]
    \centering
    \includegraphics[scale=0.6]{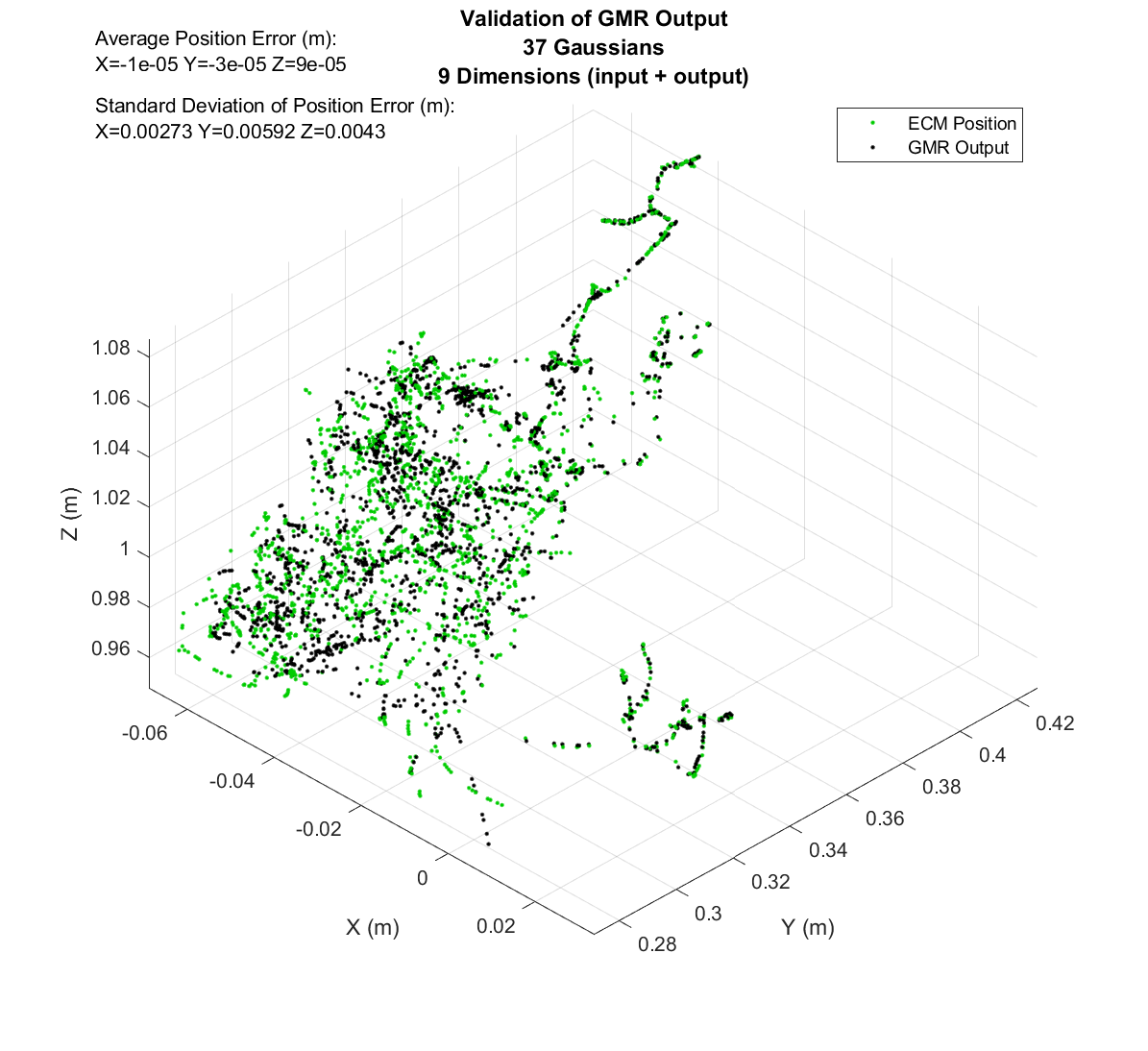}
    \caption{Validation of GMR output with position for Intuitive dataset}
    \label{fig:sinx}
\end{figure}

The first model we trained on the Intuitive Surgical dataset took only position as an input. Using this model, we found that we were able to predict the camera location with a root mean square cartesian position error of 0.1 mm and standard deviation of 4.3 mm.

\begin{figure}[H]
    \centering
    \includegraphics[scale=0.6]{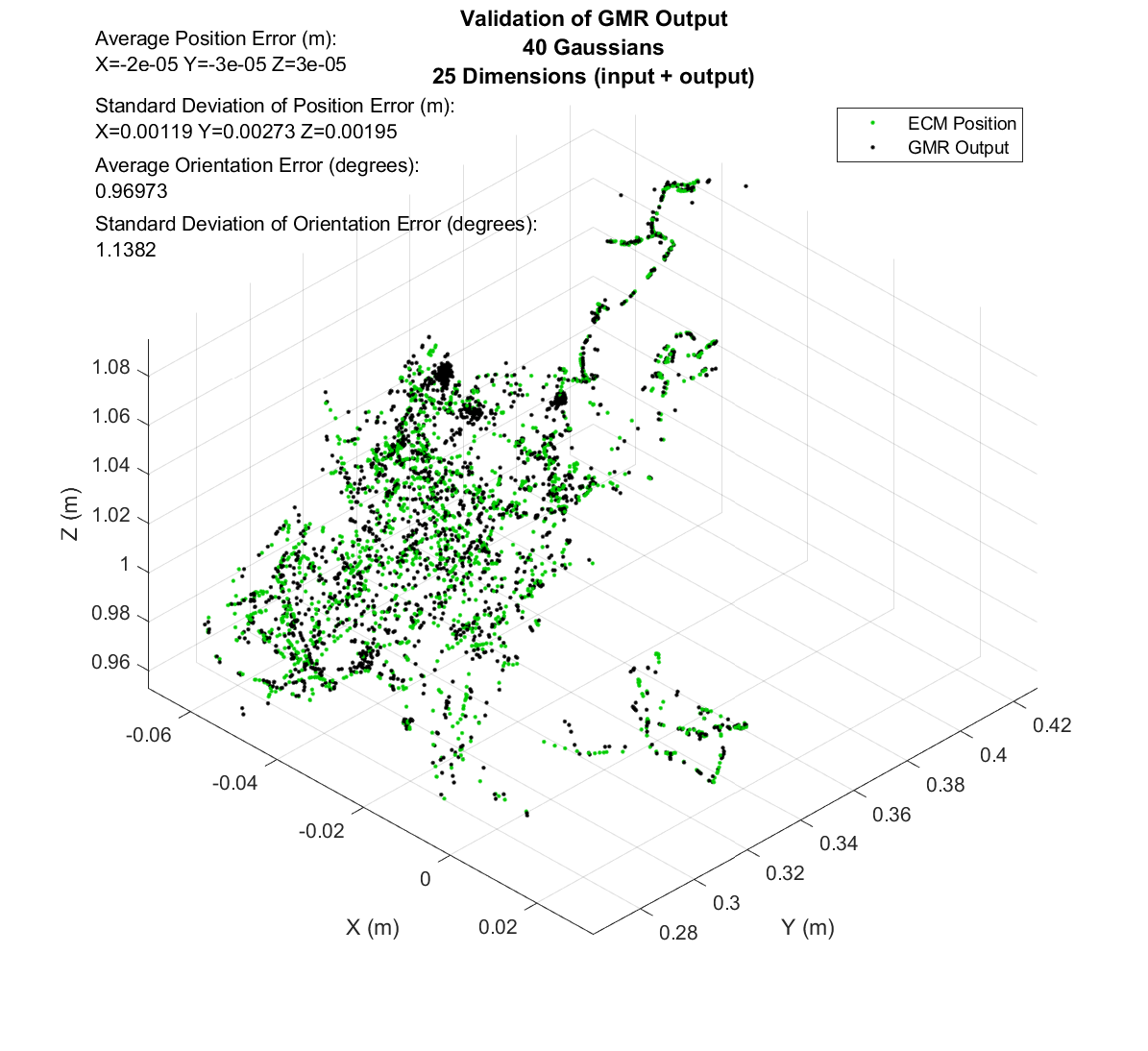}
    \caption{Validation of GMR output with position, orientation and eye gaze for Intuitive dataset}
    \label{fig:sinx}
\end{figure}

Lastly, we trained a model including position, orientation, and surgeon eye gaze as inputs. This model obtained an average position error of 0.05 mm with standard deviation 2.0 mm. For orientation predictions, this model had an average angular offset of 0.97 degrees, with standard deviation 1.14 degrees.

The following Table 4.1 summarizes the results from the two models trained on this dataset. Based on these results, we observed that adding orientation and eye gaze as inputs to the model decreased the average position error and standard deviation of error by around 50\%. 
\begin{table}[H]
        \centering
        \scalebox{0.85}{
        \begin{tabular}{|l||l|l|}
        \hline
         \textbf{Input Parameters} & \textbf{Position (mean, std) in mm} & \textbf{Orientation (mean, std) degrees}\\
        \hline
        Position & 0.1, 4.3 & N/A\\
        \hline
        Position + Orientation + eye gaze & 0.05, 2.0 & 0.97, 1.14\\
        \hline
        \end{tabular}}
        \caption{Model Training Results}
        \label{tab:Model Training Results}
    \end{table}
\chapter{Conclusions}\label{chpt:Conclusion}

In this report, we have shown that Learning from Demonstration using the da Vinci tool arm positions, orientations, and the surgeon’s eye gaze has potential to be a viable method for autonomous camera control, and it appears to offer some performance improvements over simpler models with fewer inputs. By applying our camera control system to task data from expert surgeons, we observed a 50\% decrease in prediction error when supplementing tool arm position data with orientation and eye gaze, as compared to using only tool arm positions as inputs. This means that in this setting, introducing these additional input parameters allowed us to increase the accuracy of our predictions and provide a more accurate replication of the behaviours observed in the training data. Although these results are promising, we were not able to have the model evaluated by surgeons, so we cannot definitively say whether this form of autonomous camera control will increase the surgeon’s comfort during surgical tasks. We also considered only one surgeon and one task for our models from this dataset, so we cannot be certain how well a model would generalize to other tasks or the preferences of other surgeons.

Throughout this project we also developed and tested a framework for complete data collection, a model training and deployment pipeline, and a validation mechanism for model output. With the help of this framework it would be possible to test our autonomous camera control model with the complete set of parameters (robot kinematic data and eye gaze) on the real da Vinci robot.

\chapter{Recommendations}\label{chpt:Recommendations}

Ultimately, the goal of autonomous camera control is to improve the experience for the surgeon and improve the outcome for the patient. As a next step for the project, we recommend qualitative and quantitative evaluation of our model for autonomous camera control by expert surgeons. Listed below are a number of metrics for qualitative and quantitative evaluation.

\section{Quantitative Evaluation}
\subsection{Smoothness of Camera Movement}
One of the most important features of a good camera control system is the smoothness of the camera motion. We can use dimensionless jerk (DLJ) and log dimensionless jerk (LDLJ) to estimate the smoothness of a discrete trajectory. Using these metrics, we could characterize the smoothness of the camera motion across multiple surgical tasks, which could indicate surgeon satisfaction with the model.


\subsection{Effect on Duration of Surgical Task}
  Another important metric is the time that is taken to complete a surgical task with and without the autonomous surgical camera control. This will give us a good indication if adding autonomous control has an impact of the duration of surgical procedures and provide an estimate of the efficiency of our model.
 
 \subsection{Number of Manual Overrides per Task}
    We can also measure the number of manual overrides the surgeon performs during a surgical task with the autonomous camera control. This override would be initiated by pressing the foot pedal and would revert the endoscopic camera arm to manual control. This would help us determine surgeon satisfaction with the model and learn which types of surgical tasks are better suited to autonomous camera control.
\section{Qualitative Evaluation}

For qualitative evaluation, we have the following metrics:
\begin{enumerate}
    \item Ease of Use
    \item Mental workload compared to alternate surgery method
    \item Overall surgeon satisfaction
\end{enumerate}
For all of these points, surgeons would give us a rating on the scale ranging from 1 - 5 on the use of the system with the autonomous camera control vs the system with the manual camera control.

\chapter{Deliverables}\label{chpt:Deliverables}
In our final meeting with our project sponsor, we agreed to deliver:
\begin{enumerate}
\item Documentation of code explaining how it works and how to run it.
\item Access to all the documented code used to train and validate models, generate predictions, and validate prediction output.
\item A complete framework for data collection, including robot kinematic data and surgeon eye gaze data.
\item Documentation for data processing pipeline used on the and procedure to analyze data from Intuitive Surgical
\item Code used to deploy models on the dVRK
\item Documentation of ECM reachable workspace code
\end{enumerate}

\end{mainf}

\begin{appendices}
\multappendices


\chpt{Technical Background: Robotics and Model Theory}
\section{Cartesian and Joint Space}
Robot kinematics concerns the process of commanding a robot arm to a certain pose. There are two spaces of interest in robot kinematics: cartesian space and joint space. Cartesian space is the typical coordinate space where points in space are labeled on a three dimensional grid as (x, y, z) coordinates. Labelling points in space in this fashion is intuitive and every point has a unique set of associated coordinates, making it a very popular choice of coordinate system. This contrasts joint space, which has as many dimensions as there are joints in the manipulator, and each coordinate represents the joint parameter of the corresponding joint. For revolute joints, the joint parameter is the angle, and for the prismatic joints it is the displacement from the zero position. Joint space is the most natural coordinate system for controlling a robotic manipulator, in fact required at the lowest level, as each coordinate directly corresponds to the angle each joint must be moved to to reach the desired pose.

\newpage
\section{Frame Transformations}
In robotics, the position and orientation of each end effector or joint is defined with respect to a reference frame, which is a static coordinate system from which translations and rotations are defined. We can represent each frame as a homogeneous 4x4 matrix that encapsulates the orientation and position of the end effector. 
\begin{figure}[h]
    \centering
    \includegraphics[scale=0.7]{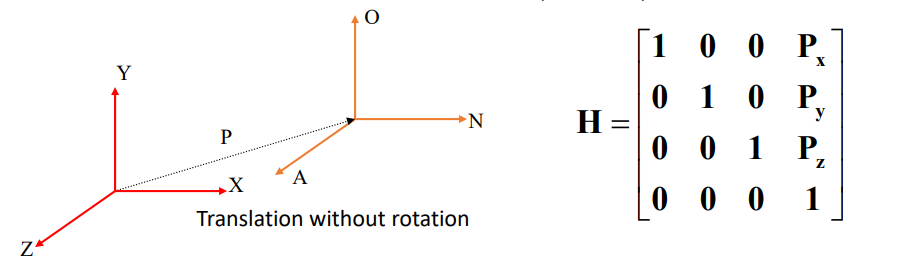}
    \caption{Homogeneous matrix, H, representing translation. \underline{\href{https://www.cs.cmu.edu/~16311/current/schedule/ppp/Lec17-FK.pdf}{Source}}}
    \label{fig:Translation}
\end{figure}

\begin{figure}[h]
    \centering
    \includegraphics[scale=0.7]{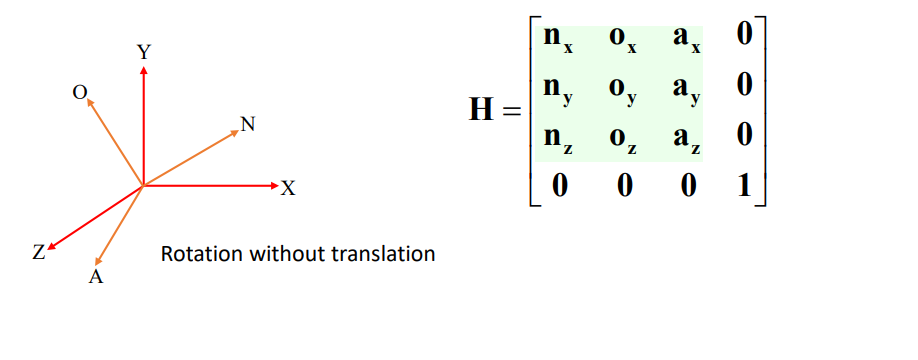}
    \caption{Homogeneous matrix, H, representing orientation. \underline{\href{https://www.cs.cmu.edu/~16311/current/schedule/ppp/Lec17-FK.pdf}{Source}}}
    \label{fig:Orientation}
\end{figure}

\newpage We can combine the rotation and translation into a single homogeneous 4x4 matrix if and only if both the rotation and position are defined with respect to the same point.
\begin{figure}[H]
    \centering
    \includegraphics[scale=0.3]{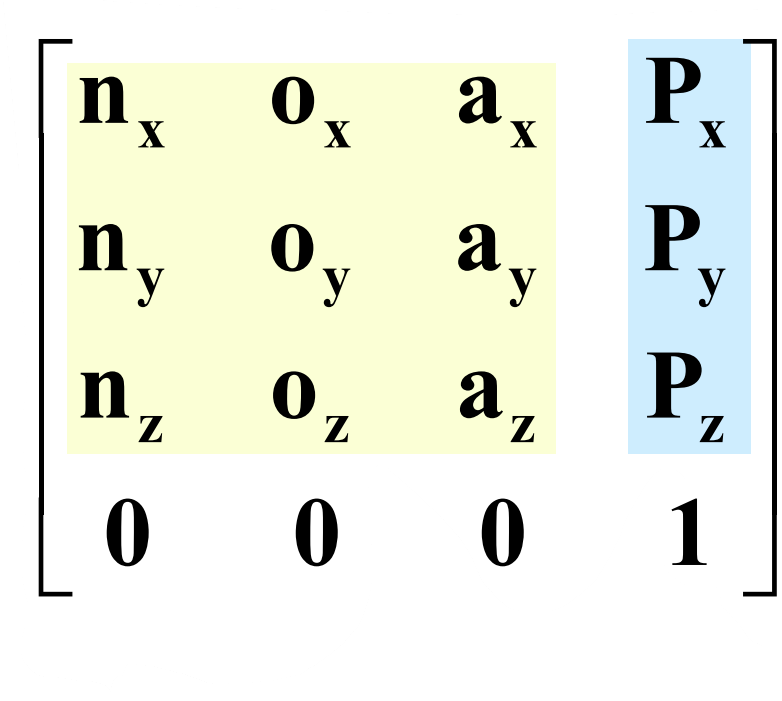}
    \caption{Homogeneous matrix, H, representing position and orientation. \underline{\href{https://www.cs.cmu.edu/~16311/current/schedule/ppp/Lec17-FK.pdf}{Source}}}
    \label{fig:Position and Orientation}
\end{figure}
This 4x4 matrix representation allows us to go from any particular reference frame to our end position of interest, given we know the subsequent 4x4 homogeneous matrices. 

For example in Figure~\ref{fig:CoordinateFrame4}, if we want to define the point w, which is originally defined in the base coordinate system of (N, O, A),  with respect to the base frame (X,Y,Z), we can do this by simple matrix multiplication of the internal base frames:
\begin{figure}[H]
    \centering
    \includegraphics[scale=0.7]{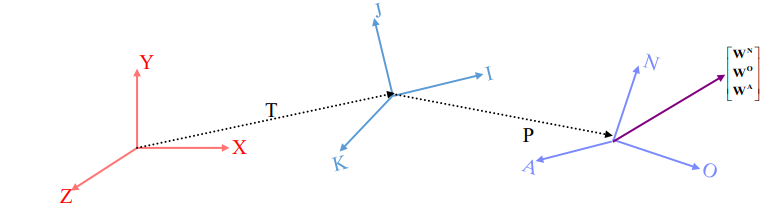}
    \caption{Transformation chain from world coordinate to point \textbf{W}. \underline{\href{https://www.cs.cmu.edu/~16311/current/schedule/ppp/Lec17-FK.pdf}{Source}}}
    \label{fig:CoordinateFrame4}
\end{figure}
\begin{figure}[H]
    \centering
    \includegraphics[scale=0.7]{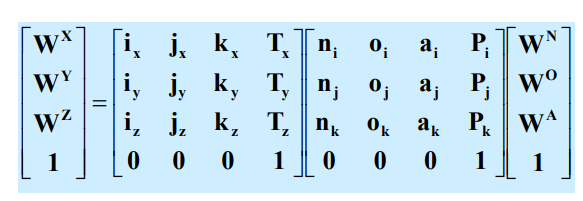}
    \caption{Transformation chain from world coordinate to point \textbf{W}. \underline{\href{https://www.cs.cmu.edu/~16311/current/schedule/ppp/Lec17-FK.pdf}{Source}} }
    \label{fig:CoordinateFrame5}
\end{figure}

\subsection{Frame Transformations in da Vinci Robot}

Each da Vinci Patient-Side Manipulator (PSM) is supported by Setup Joints (SUJ) that connect it to the central robot shaft. These SUJs position the PSMs at pre-planned incision ports and remain stationary. The point of incision serves as a remote center of motion, and the tooltip moves about this center. Thus, the SUJs essentially fix the base of each arm. The main coordinate frames to consider are shown in Fig. 2.6. The endoscope is held by the Endoscope Control Manipulator (ECM). 
\begin{figure}[H]
    \centering
    \includegraphics[scale=0.7]{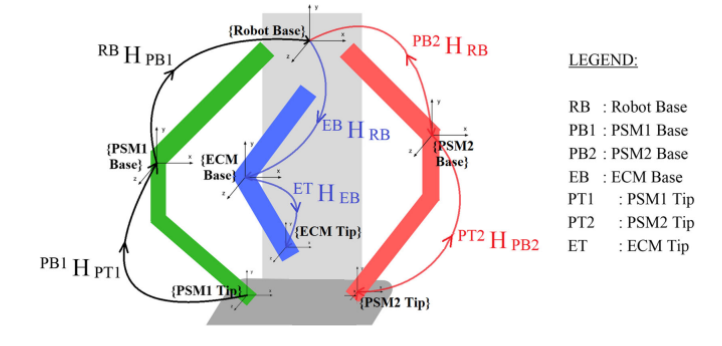}
    \caption{Relevant coordinate systems of the da Vinci and the relevant transformations. The transformation with respect to ECM is shown
in black and blue paths. \underline{\href{https://people.ece.ubc.ca/~aabdelaal/index.html/IJCARS\%202019.pdf}{Source}}}
    \label{fig:Block diagram}
\end{figure}
To achieve visual-motor consistency (hand-eye coordination), the end effectors of the PSMs need to be defined with respect to the ECM tip. The transformation \begin{equation}
    {}^{X}\textbf{H}_{Y}
\end{equation} represents the transformation from frame X to frame Y.

The black path continued by the blue path shows the position of PSM1 tip with respect to the ECM tip. More formally, we can write this as the following equation:
\begin{equation}
    {}^{ET}\textbf{H}_{PT1} = {}^{ET}\textbf{H}_{EB} {}^{EB}\textbf{H}_{RB} {}^{RB}\textbf{H}_{PB1} {}^{PB1}\textbf{H}_{PT1}
\end{equation} represents the transformation from ECM tip to PSM1 tip.

Similarly the red path followed by the blue path shows the position of PSM2 tip with respect to the ECM tip. The equation is given as follows:
\begin{equation}
    {}^{ET}\textbf{H}_{PT2} = {}^{ET}\textbf{H}_{EB} {}^{EB}\textbf{H}_{RB} {}^{RB}\textbf{H}_{PB2} {}^{PB2}\textbf{H}_{PT2}
\end{equation} represents the transformation from ECM tip to PSM2 tip.
The positions of the tool tip (defined with respect to
the PSM Base) are transformed to the ECM frame by using
these transformations. This way, all movements of the tool
tip will be in the frame of the endoscope, thus maintaining
visual-motor consistency when looking at the images from
the endoscope.

\subsection{Frame Transformations to World Coordinate Frame}
\label{Frame Transform to WC}
During data collection (explained later in Section \ref{data Collection System}) we need to ensure that the PSM tool tips and ECM tip are all defined with respect to a global base frame. Referring to Fig. 2.6, this would be obtained using the following transformations:
\begin{equation}
    {}^{RB}\textbf{H}_{PT1} = {}^{RB}\textbf{H}_{PB1} {}^{PB1}\textbf{H}_{PT1} 
\end{equation}
\begin{equation}
    {}^{RB}\textbf{H}_{ET} = {}^{EB}\textbf{H}_{RB} {}^{ET}\textbf{H}_{EB} 
\end{equation}
\begin{equation}
    {}^{RB}\textbf{H}_{PT2} = {}^{RB}\textbf{H}_{PB2} {}^{PB2}\textbf{H}_{PT2} 
\end{equation}

\section{Forward and Inverse Kinematics}
As cartesian space is the most convenient coordinate system for controlling a robot manipulator at a higher level, and joint space is required at the lowest level, we must have a way to convert between each coordinate system. Calculating the cartesian pose from a given point in joint space is called “Forward Kinematics” and is a relatively straightforward process as each point in joint space maps to exactly one point in cartesian space. Going in the opposite direction, calculating the joint space coordinates from a given pose in cartesian space, is called “Inverse Kinematics” and is an open field of research in robotics. In general, the inverse kinematics of a given robotic manipulator is a very difficult problem to solve, often having multiple solutions. Moreover, most methods for finding the inverse kinematics, called inverse kinematic solvers, fail around points called singularities.

\section{Reachable Workspace and Singularities} \label{reachableWS}

In robotics, a singularity is a joint configuration where the end effector’s movement loses a degree of freedom (DoF). This often occurs when the axes of two (or more) joints align, resulting in the same motion of the end effector when each joint moves. A related idea is that of the reachable workspace of a robotic manipulator. The reachable workspace is defined as all positions the end effector is able to move to, irrespective of its orientation, within the joint limits. To state this more precisely, if we let $\boldsymbol{x}$ be a position in cartesian space, $\boldsymbol{q}$ a vector in joint space, $\boldsymbol{S}$ the joint space, and $\boldsymbol{o_n(q)}$ the position of the end effector in cartesian space as a function of the joint parameters, then the reachable workspace is defined as $\{ \boldsymbol{x} | \exists \boldsymbol{q} \in \boldsymbol{S} \text{ such that } \boldsymbol{o_n(q)} = \boldsymbol{x} \}$ \citep{tims}

\section{Denavit-Hartenberg Parameters} \label{DH}

The Denavit-Hartenberg (DH) convention is a popular approach for placing reference frames and transforming between them. In the DH convention reference frames are placed at the joints of a robot manipulator, and the DH parameters are four parameters which completely describe the translations and rotations between adjacent reference frames. 

There are two conventions within the DH convention which differ in how reference frames are attached and the order of transformations performed to change between frames. The two conventions are the classic or distal convention, and the prismatic or modified convention. The dVRK uses the modified convention, described below.

Reference frames are attached to joints, labelled in ascending order such that frame 0 is the frame furthest from the tip of the robot manipulator. For any given frame, the z-axis is aligned along the axis of the associated joint, the x-axis points towards and intersects the z-axis of the next frame, and the y-axis is placed to complete a right-handed coordinate system. 

For frame $i-1$, the (modified) DH parameters are given in Table~\ref{tab:DH-def}.

\begin{table}[H]
    \centering
    \begin{tabular}{r|l}
        \textbf{Symbol} & \textbf{Description}\\
        \hline
        $d_i$ & Translation along current z-axis to new origin\\
        $\theta_i$ & Angle about current z-axis from current x-axis to new x-axis\\
        $a_i$ & Translation along current x-axis to new origin\\
        $\alpha_i$ & Angle about current x-axis from current z-axis to new z-axis
    \end{tabular}
    \caption{Descriptions of the Modified DH parameters for a Transform from Frame $i-1$ to Frame $i$.}
    \label{tab:DH-def}
\end{table}

\begin{figure}
    \centering
    \includegraphics[scale=0.7]{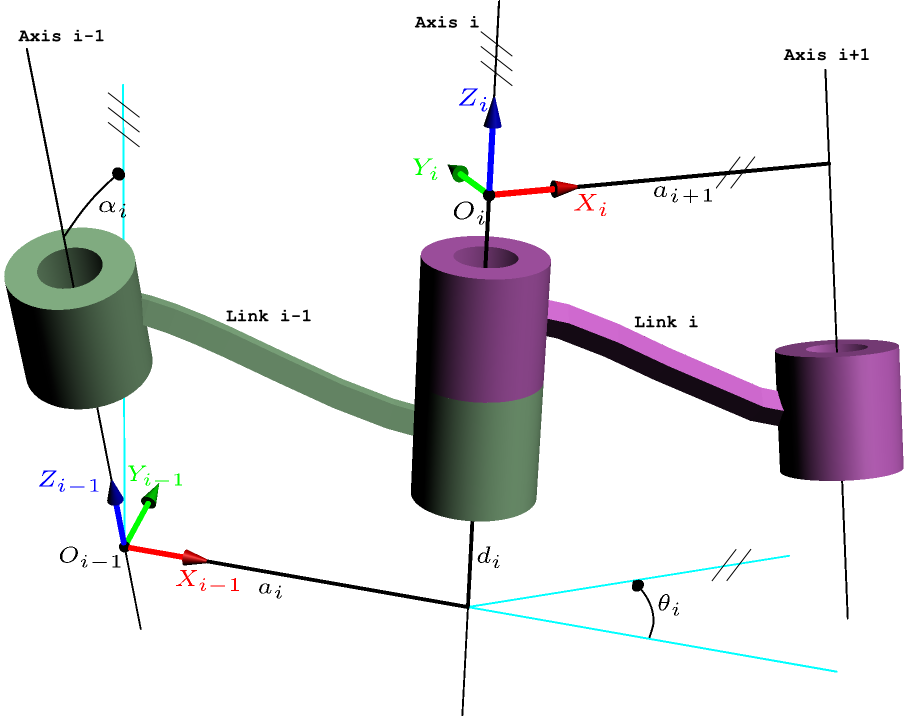}
    \caption{Modified DH parameters. \underline{\href{https://upload.wikimedia.org/wikipedia/commons/d/d8/DHParameter.png}{Source}}}
    \label{fig:Appendix_DH}
\end{figure}

Figure~\ref{fig:Appendix_DH} illustrates these frames and DH parameters in a general case.

The transform from frame $i-1$ to frame $i$ is given by the following homogeneous 4x4 transform matrix:

\begin{equation}
    {}^{i-1}T_{i} = \left[ \begin{array}{ccc|c}
    \cos\theta_i & -\sin\theta_i  & 0 & a_i \\
    \sin\theta_i \cos\alpha_i & \cos\theta_i \cos\alpha_i & -\sin\alpha_i & -d_i \sin\alpha_i \\
    \sin\theta_i\sin\alpha_i & \cos\theta_i \sin\alpha_i & \cos\alpha_i & d_i \cos\alpha_i \\
    \hline
    0 & 0 & 0 & 1
    \end{array} \right]
\end{equation}

The transform matrix from frame 0 to the end effector frame n can be obtained by composing the transformations, or more precisely,

\begin{equation}
    {}^{0}T_{n} = {}^{0}T_{1} {}^{1}T_{2} ... {}^{n-1}T_{n} = \prod_{j=1}^{n} {}^{j-1}T_{j}
\end{equation}

\chpt{ECM Reachable Workspace Calculations} \label{workspace_calculations}

To find the reachable workspace of the ECM, we use the ECM’s DH parameters and joint limits, listed in Tables~\ref{tab:ECM-DH-param} and \ref{tab:ECM-joint-limits} below.

\begin{table}[H]
    \centering
    \begin{tabular}{c|c|c|c|c|c}
        \textbf{Frame} & \textbf{Joint Type} & $d$ & $\theta$ & $a$ & $\alpha$\\
        \hline
        1 & Revolute & $0$ & $\frac{\pi}{2}$ & $0$ & $\frac{\pi}{2}$\\
        2 & Revolute & $0$ & $-\frac{\pi}{2}$ & $0$ & $-\frac{\pi}{2}$\\
        3 & Prismatic & $-0.3822$ & $0$ & $0$ & $\frac{\pi}{2}$\\
        4 & Revolute & $0.3829$ & $0$ & $0$ & $0$
    \end{tabular}
    \caption{ECM modified DH parameters}
    \label{tab:ECM-DH-param}
\end{table}

\begin{table}[H]
    \centering
    \begin{tabular}{c|c|c|c}
        \textbf{Joint} & \textbf{Unit} & \textbf{Lower Limit} & \textbf{Upper Limit}\\
        \hline
        1 & Radians & $-1.5708$ & $1.5708$\\
        2 & Radians & $-0.7853$ & $1.1344$\\
        3 & Meters & $0.0$ & $0.235$\\
        4 & Radians & $-1.5708$ & $1.5708$
    \end{tabular}
    \caption{ECM joint limits}
    \label{tab:ECM-joint-limits}
\end{table}

Using the DH parameters we construct the 4x4 homogeneous transformation matrices ${}^{i-1}T_{i}$ describing the transform between each adjacent reference frame and calculate the transformation from the RCM frame to the end effector frame as $T = {}^0T_1 {}^1T_2 {}^2T_3 {}^3T_4$. The entries of the matrix $T$ are given below, where $q_i$ is the joint parameter for joint $i, i \in \{ 1,2,3,4 \}$.

\begin{equation}
\begin{split}
    T_{11} &= - \cos(q_1)\sin(q_4) - \cos(q_4)\sin(q_1)\sin(q_2)\\
    T_{12} &= \sin(q_1)\sin(q_2)\sin(q_4) - \cos(q_1)\cos(q_4)\\
    T_{13} &= \cos(q_2)\sin(q_1)\\
    T_{14} &= \cos(q_2)\sin(q_1)(q_3 + 7/10000)\\
    T_{21} &= - \cos(q_2)\cos(q_4)\\
    T_{22} &= \cos(q_2)\sin(q_4)\\
    T_{23} &= -\sin(q_2)\\
    T_{24} &= -\sin(q_2)(q_3 + 7/10000)\\
    T_{31} &= \cos(q_1)\cos(q_4)\sin(q_2) - \sin(q_1)\sin(q_4)\\
    T_{32} &= - \cos(q_4)\sin(q_1) - \cos(q_1)\sin(q_2)\sin(q_4)\\
    T_{33} &= -\cos(q_1)\cos(q_2)\\
    T_{34} &= -\cos(q_1)\cos(q_2)(q_3 + 7/10000)\\
    T_{41} &= 0\\
    T_{42} &= 0\\
    T_{43} &= 0\\
    T_{44} &= 1
\end{split}
\end{equation}

\begin{figure}[H]
    \centering
    \includegraphics[scale=0.7]{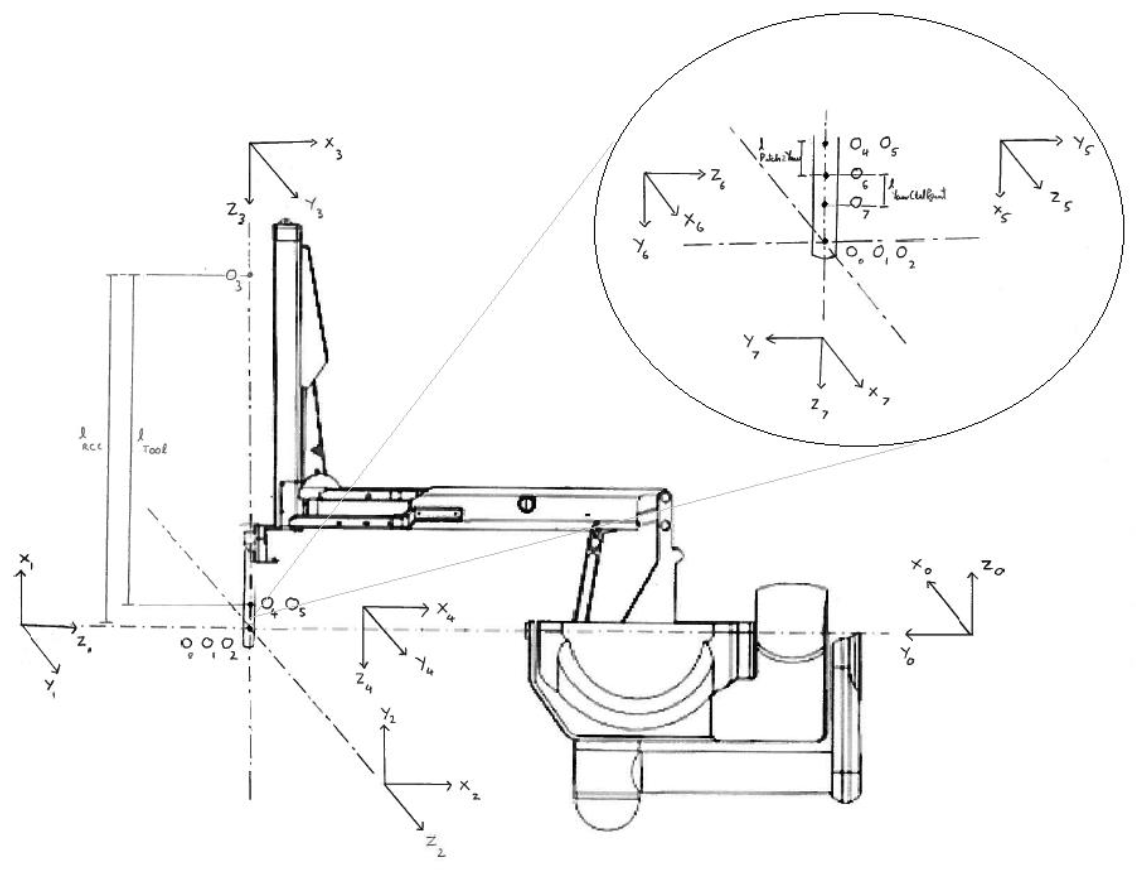}
    \caption{ECM frames}
    \medskip
    \small The reference frames associated with the (modified) DH parameters for the ECM. Frames 5, 6, and 7 shown in the bubble are the wrist frames for the PSMs, which are shown since the ECM and PSMs share the same first five reference frames, 0 through 4.
    \label{fig:ECM-frames}
\end{figure}

Figure~\ref{fig:ECM-frames} shows the ECM overlaid with the location and orientation of each intermediate frame, as well as frames 0 and 4, which are the RCM frame and end effector frame respectively.

The elements of T represent the position and orientation of the end effector with respect to the RCM frame, so we have:

\begin{equation}
    \boldsymbol{p} = T[1:3,4]
\end{equation}

\begin{equation}
        \boldsymbol{R} = T[1:3,1:3]
\end{equation}

$\boldsymbol{p}$ and $\boldsymbol{R}$ constitute the forward kinematics equations. As expected from the geometry of the ECM, we see that the fourth joint angle only affects the orientation, and the third joint angle only affects the position. Cognisant of the joint limits, the inverse kinematics can be solved exactly, taking $U_1$ to be the upper limit of joint 1, as:

\begin{equation}
    q_1 = \begin{cases} 
      \atan \left ( -\frac{\boldsymbol{p}(1)}{\boldsymbol{p}(3)} \right ) & \text{if } \boldsymbol{p}(3) \neq 0 \\
      \sgn (\boldsymbol{p}(1))*U_1 & \text{if } \boldsymbol{p}(1) \neq 0 \cap \boldsymbol{p}(3) = 0
   \end{cases}
\end{equation}

\begin{equation}
    q_2 = \begin{cases} 
      \atan \left ( \frac{\boldsymbol{p}(2)}{\boldsymbol{p}(3)} \cdot \cos(q_1) \right ) & \text{if } \boldsymbol{p}(3) \neq 0 \\
      \atan \left ( -\frac{\boldsymbol{p}(2)}{\boldsymbol{p}(1)} \cdot \sin(q_1) \right ) & \text{if } \boldsymbol{p}(1) \neq 0 \cap \boldsymbol{p}(3) = 0
   \end{cases}
\end{equation}

\begin{equation}
    q_3 = |\boldsymbol{p}| - \frac{7}{10000}
\end{equation}

\begin{equation}
    q_4 = \atan2(\boldsymbol{R}(2,2),-\boldsymbol{R}(2,1))
\end{equation}

\begin{figure}[H]
    \centering
    \includegraphics[width=0.9\columnwidth]{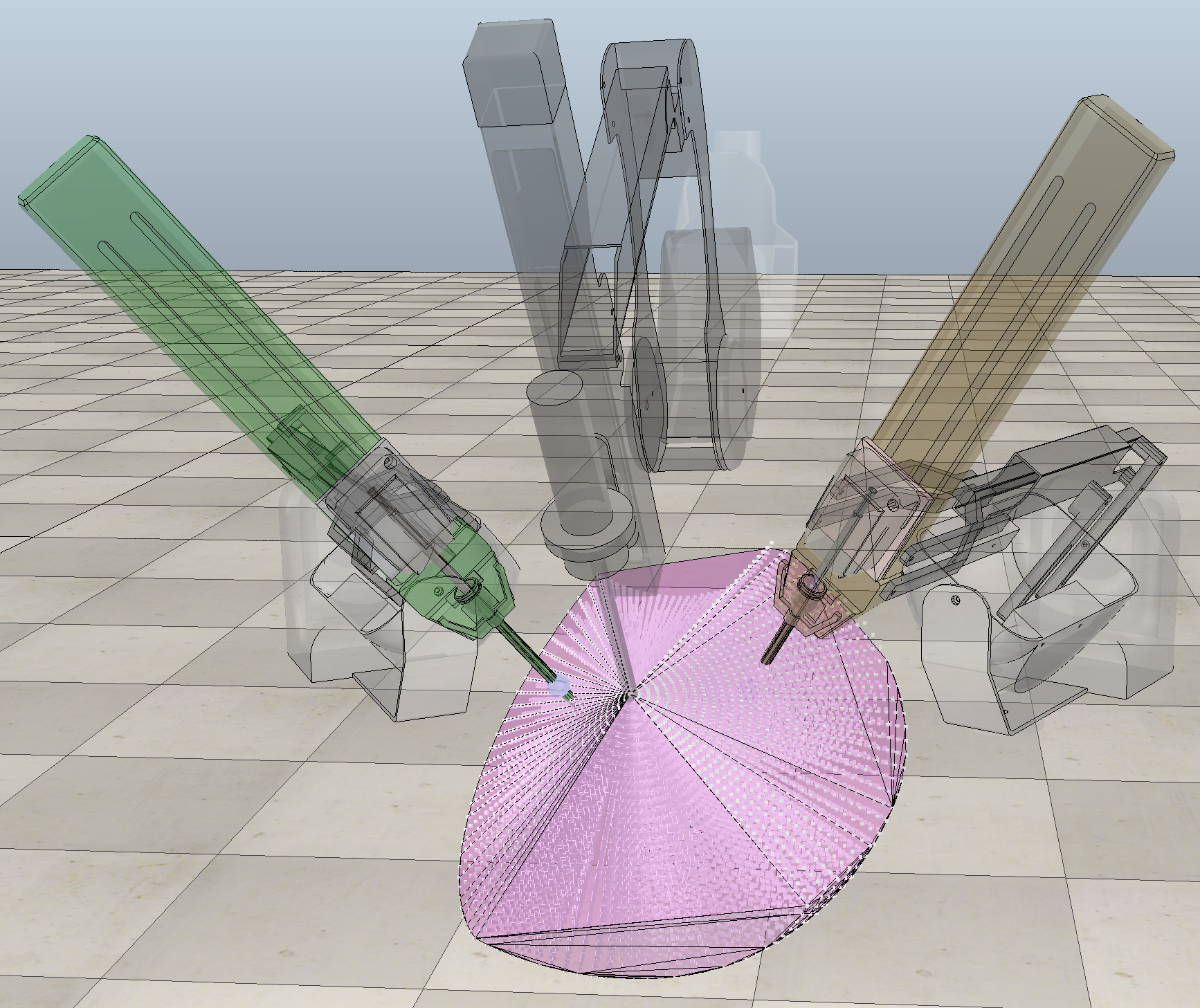}
    \caption{CoppeliaSim workspace}
    \medskip
    \small A convex hull (pink) of the reachable workspace of the ECM (central, black manipulator). The points visited to outline the workspace are visible throughout the convex hull as white spheres. The two bounding cones and two bounding planes are clear, the surface at the central point is a small sphere, and the far side surface is a larger sphere.
    \label{fig:copsim-workspace}
\end{figure}

Any cases not covered by the conditions for joints 1 and 2 represent points that are outside of the reachable workspace. Using the forward kinematics equations and the joint limits, we find that the workspace is a solid volume bounded by two spheres, two cones, and two planes. This calculation was qualitatively verified using CoppeliaSim, where by visiting a large number of points and placing a dot at each location, a discrete approximation of the workspace formed which resembled what was expected from the calculations. Figure~\ref{fig:copsim-workspace} shows this iteratively-generated workspace in CoppeliaSim.

\chpt{Hardware and Software}
\label{Hardware and Software}
\section{da Vinci Research Kit (dVRK)}
To obtain the position and orientation data from the da Vinci surgical system, we used the da Vinci Research Kit (dVRK), an open source ROS based control system developed at Johns Hopkins University. The dVRK consists of electronic controllers and firmware that provides a real-time interface with the da Vinci robot.
For each robotic “arm” (slave arm, master manipulator, endoscope manipulator, there is one controller box used to interface with the arm (see Figure 2.5) In addition, a setup-joint controller box is used which enables to configure the chain of transformation between the arms.
The dVRK has controller boxes which as an electronic interface between a user and the da Vinci robot.
With two slave arms, two master manipulators, one setup-joint controller, and one endoscope manipulator, the dVRK has 6 controller boxes. Within each controller box are two sets of electronic boards. Each controller box has a Quad Linear Amplifier (QLA) board and an IEEE-1394 Field-Programmable Gate Array (FPGA) Controller board. The QLA board drives
four DC motors, and has Analog-to-Digital (ADC) converters as well as Digital-
to-Analog (DAC) converters for setting and monitoring the motor current. The
FPGA Controller board controls the motors, and communicates with the an external Linux computer with the FireWire protocol. Closed loop control of the robotic
arm is achieved with sensor feedback connected to the FPGA board.
\begin{figure}[H]
    \centering
    \includegraphics[scale=0.7]{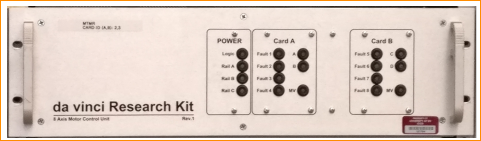}
    \caption{dVRK controller box. \underline{\href{https://github.com/jhu-dvrk/sawIntuitiveResearchKit/wiki}{Source}}}
    \label{fig:Block diagram}
\end{figure}
\begin{figure}[H]
    \centering
    \includegraphics[scale=0.4]{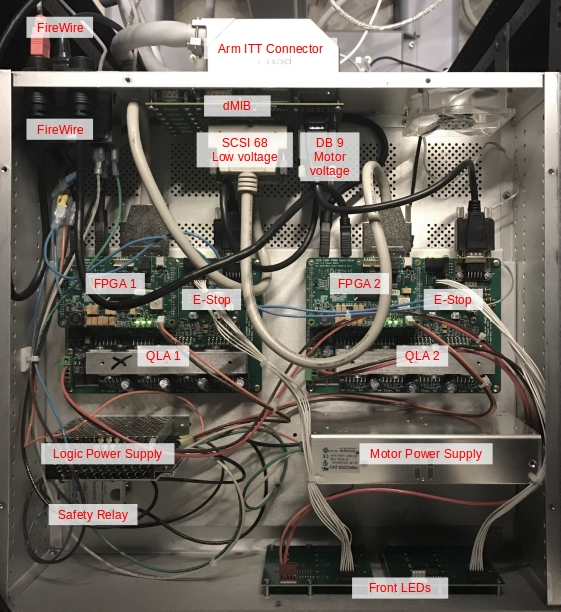}
    \caption{dVRK controller box internal components. \underline{\href{https://github.com/jhu-dvrk/sawIntuitiveResearchKit/wiki}{Source}}}
    \label{fig:Block diagram}
\end{figure}
The software for the DVRK is built upon the Center for Computer-Integrated
Surgical Systems and Technology (“cisst”) package and Surgical Assistant Work-
station (SAW) open-source libraries developed at the Johns Hopkins University
(JHU). The cisst package (named after the Center for Computer-Integrated Surgical Systems and Technology at JHU) contains low level functionality for computer-assisted interventions. The SAW libraries contain robotic and imaging functionality for computer-assisted surgical applications. The overall underlying software architecture for dVRK is shown below:
\begin{figure}[H]
    \centering
    \includegraphics[scale=0.3]{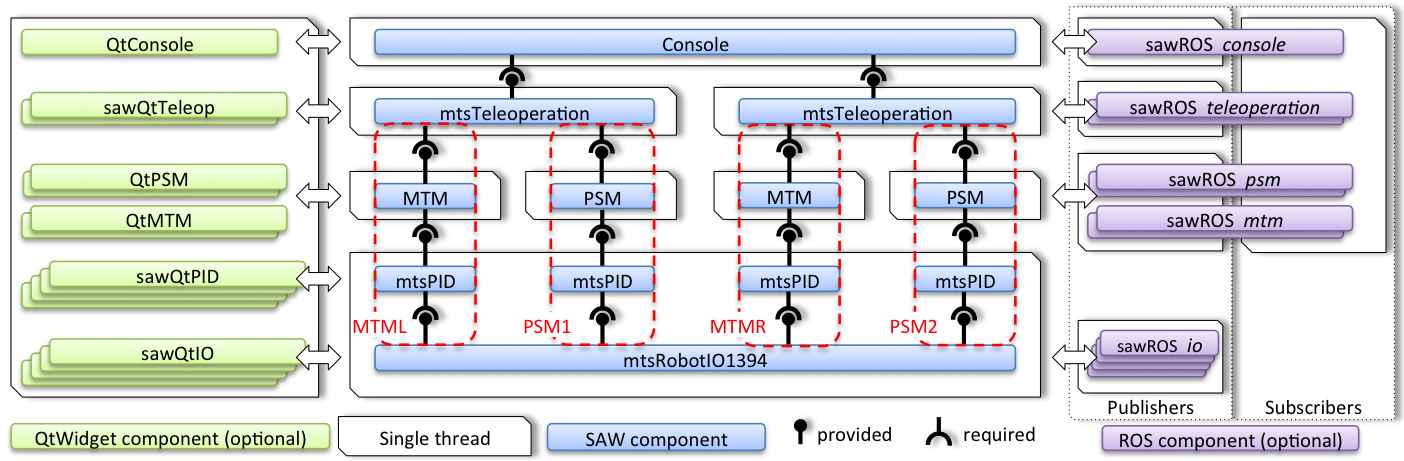}
    \caption{dVRK software flow diagram. \underline{\href{https://github.com/jhu-dvrk/sawIntuitiveResearchKit/wiki}{Source}}}
    \label{fig:Block diagram}
\end{figure}
A console GUI application is provided which contains an interface for sensor monitoring, PID control, and tele-operation status. 

An example of the GUI is shown below:
\begin{figure}[H]
    \centering
    \includegraphics[scale=0.18]{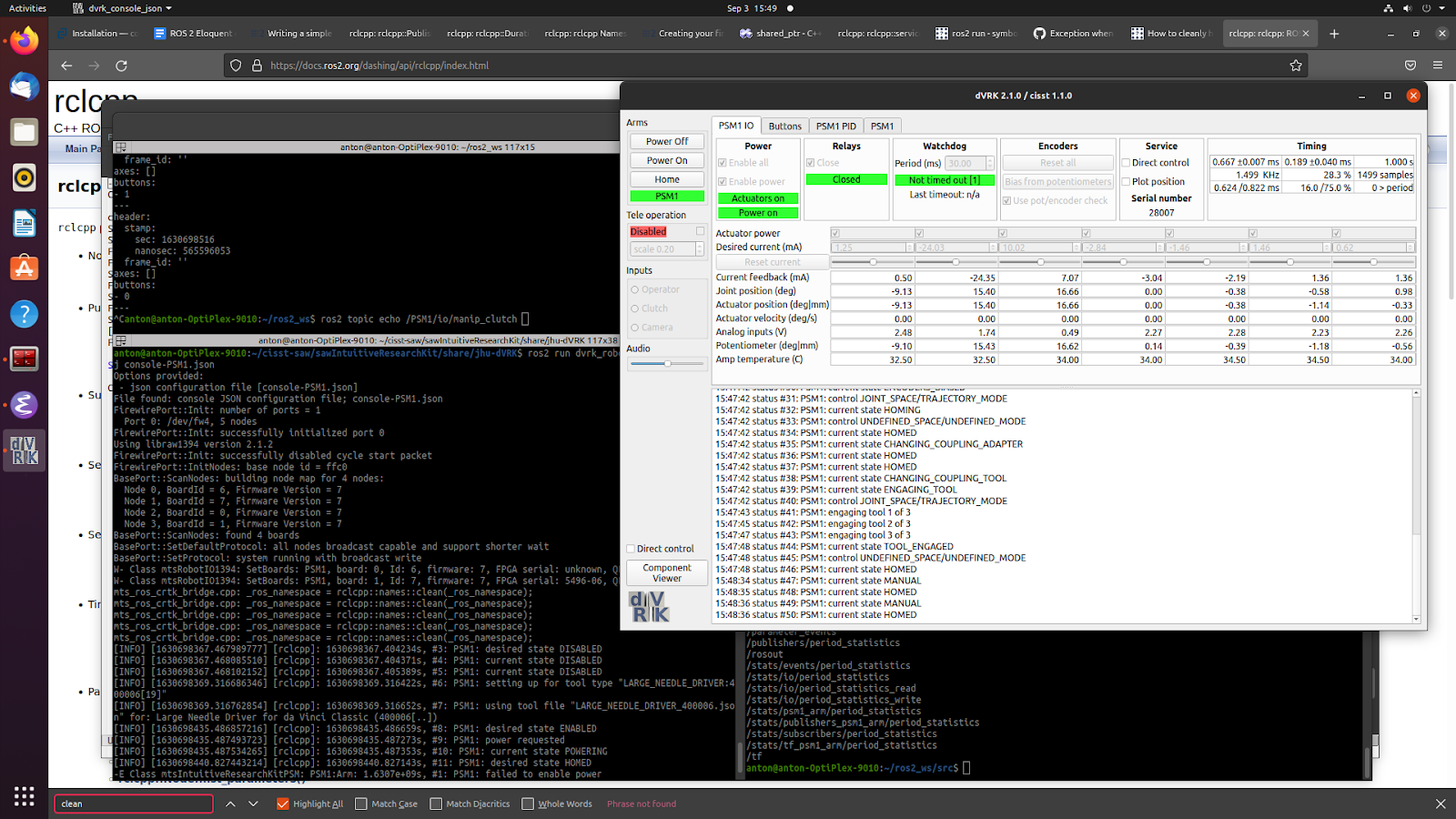}
    \caption{dVRK console GUI. \underline{\href{https://github.com/jhu-dvrk/sawIntuitiveResearchKit/wiki}{Source}}}
    \label{fig:Block diagram}
\end{figure}
All code for underlying robot control is written in C++ with Python wrappers using the ROS framework. dVRK enables ROS communication via “ROS topics” that allows information to be easily read and sent to the da Vinci system. The PC used is a Linux-based system with Ubuntu 18.04, ROS Melodic, and dVRK v2.10.

\chpt{Model Theory}

\section{Clustering}
Clustering is an action of grouping a collection of data points such that data points with more similarities are in the same group or cluster. The goal of clustering is to segregate a number of data points with similar traits and assign them into clusters. Clustering allows us to find patterns in high dimensional data, which is essential in learning from demonstration. 
\section{K-Means Clustering Method}
One of the clustering methods, K-Means, minimizes the squared Euclidean distance of the target data point and the centroid of the clusters. The “mean” in K-Means indicates that the algorithm intends to find the “average” of the data points. The K-Means algorithm defines k number of centroids and allocate data points to their nearest clusters. It is worth noting that these centroids are discrete centroids, i.e., plotting those k centroids would be a scatter plot in the n-d space for n-dimensional data. More importantly, K-Means is a Hard Clustering Method, which means that it assigns each data point to one and only one centroid. This will be a limitation in our project

\chpt{Pipeline}
\section{Deployment Pipeline}
There are multiple setups and transformations that the davinci robot could operate in, so it is crucial that we collect the data following these steps to ensure the validity of the data. The python scripts for collecting data, generating trajectories, and replaying data on both the simulation and the real robot have been completed. Future researchers will be able to easily reproduce the results and generate more data using this pipeline. 

\begin{figure}[H]
    \centering
    \includegraphics[scale=0.64]{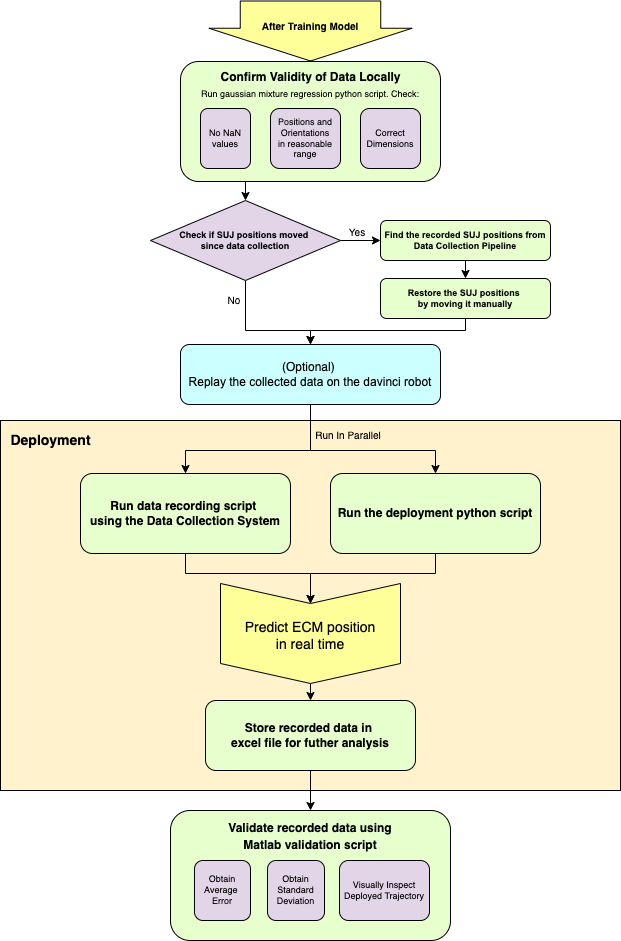}
    \caption{Detailed flowchart for deployment pipeline}
    \label{fig:pipelinedeploy}
\end{figure}

\end{appendices}

\begin{bibliof}
\bibliography{bibliography}
\end{bibliof}
\end{document}